\documentclass[lettersize,journal]{IEEEtran}
\usepackage{amsmath,amsfonts}
\usepackage{algorithmic}
\usepackage{algorithm}
\usepackage{array}
\usepackage[caption=false,font=normalsize,labelfont=sf,textfont=sf]{subfig}
\usepackage{textcomp}
\usepackage{stfloats}
\usepackage{url}
\usepackage{verbatim}
\usepackage{graphicx}
\usepackage{cite}
\usepackage[table,xcdraw]{xcolor}
\usepackage{colortbl}
\usepackage{booktabs}
\usepackage{xcolor}
\usepackage{orcidlink} 

\begin{document}

\title{ResPrune: Text-Conditioned Subspace Reconstruction for Visual Token Pruning in Large Vision-Language Models}

\author{Xu Li$^{\orcidlink{0009-0001-2431-3410}}$, Yi Zheng$^{\orcidlink{0009-0006-1549-6979}}$, Yuxuan Liang$^{\orcidlink{0009-0004-5039-7568}}$, Zhe Liu$^{\orcidlink{0009-0003-3668-6518}}$, Xiaolei Chen$^{\orcidlink{0009-0005-5700-9326}}$,  \\ Haotian Chen$^{\orcidlink{0009-0001-0593-5281}}$,  Rui Zhu$^{\orcidlink{0009-0007-7226-6923}}$, Xiangyang Xue\textsuperscript{$\dagger$}$^{\orcidlink{0000-0002-4897-9209}}$, Member, IEEE
\thanks{This work is supported by Shanghai Key Laboratory of Intelligent Information Processing and College of Computer Science and Artificial Intelligence, Fudan University. \textbf{$\dagger$denotes the corresponding author}.}
\thanks{ X. Li,Y. Zheng, Y. Liang, Z. Liu, X. Chen, H. Chen,  R. Zhu, and X. Xue are with the College of Computer Science and Artificial Intelligence, Fudan University, Shanghai 200433, China (E-mails: {xu\_li23, yxliang25,zhengy23,zheliu24,  chenxl23, htchen24,   rzhu24}@m.fudan.edu.cn, xyxue@fudan.edu.cn)}}


\markboth{Journal of \LaTeX\ Class Files,~Vol.~14, No.~8, August~2021}%
{Shell \MakeLowercase{\textit{et al.}}: A Sample Article Using IEEEtran.cls for IEEE Journals}


\maketitle


\begin{abstract}
Large Vision-Language Models (LVLMs) rely on dense visual tokens to capture fine-grained visual information, but processing all these tokens incurs substantial computational and memory overhead during inference.
To address this issue, we propose \textbf{ResPrune}, a training-free visual token pruning framework that enables efficient LVLM inference by selecting a compact yet informative subset of visual tokens.
ResPrune formulates visual token pruning as a subspace reconstruction problem and employs a greedy subspace expansion strategy guided by residual energy, allowing it to preserve the geometric structure of the original visual token space. 
To further incorporate crossmodal alignment, the selection process is conditioned on textual relevance, encouraging the retention of tokens that are both informative and instruction-relevant.
The proposed method is lightweight and model-agnostic, and can be seamlessly integrated into existing LVLM pipelines without retraining or architectural modifications.
Extensive experiments on multiple LVLM backbones, including LLaVA-1.5, LLaVA-NeXT, and Qwen2.5-VL, demonstrate that ResPrune consistently outperforms existing pruning approaches across a wide range of benchmarks, while achieving effective reductions in computation, memory consumption, and inference latency.
\end{abstract}

\begin{IEEEkeywords}
Large Vision-Language Models, Multimodal Large Language Models, Visual Token Pruning.
\end{IEEEkeywords}

\section{Introduction}
\IEEEPARstart{B}{y} integrating pre-trained vision encoders with Large Language Models (LLMs), Large Vision–Language Models (LVLMs) have emerged as a powerful paradigm for unified multimodal understanding \cite{llava, llava1.5, llavanext, instructblip}, enabling a wide range of tasks such as visual question answering \cite{tcsvt_vqa}, image captioning \cite{tcsvt_image_caption}, and multimodal reasoning \cite{tcsvt_mm_reasoning}. With the rapid advancement of visual backbones toward higher resolution and finer-grained representations, modern LVLMs typically encode an input image into hundreds to thousands of visual tokens that are processed by the LLM \cite{qwen2vl}. While such dense tokenization improves visual perception, it imposes a heavy token load on the language model, resulting in substantial computational and memory overhead and severely limiting the applicability of LVLMs in resource-constrained or latency-sensitive scenarios.

Despite the large number of visual tokens, many of them have been shown to be redundant in LVLMs \cite{fastervlm}. 
Therefore, visual token pruning has become an active research direction in the LVLM community.
Existing approaches can be broadly categorized into three groups according to the signals they exploit for token selection. 
\textbf{Visual attention–based methods} leverage attention patterns within the vision encoder \cite{prumerge, visionzip, vispruner}. When a CLS token is available, these methods typically retain tokens with high CLS-to-patch attention scores; otherwise, they rely on aggregated patch-to-patch attention. Such approaches are motivated by the assumption that visually salient regions are the most informative. 
\textbf{Crossmodal attention–based methods} instead exploit text-to-visual attention extracted from shallow layers of the LLM~\cite{fastv, sparsevlm, pdrop}. By selecting visual tokens that receive strong attention from textual tokens, these methods aim to preserve tokens that are most relevant to the input instruction.
\textbf{Diversity-based methods} take a different perspective by explicitly modeling redundancy among visual tokens \cite{divprune, dart, scope}. They measure pairwise similarity in the visual embedding space and select a subset that maximizes diversity, thereby encouraging broader coverage of the visual content.

\begin{figure}[t]
  \centering
  \includegraphics[width=\linewidth]{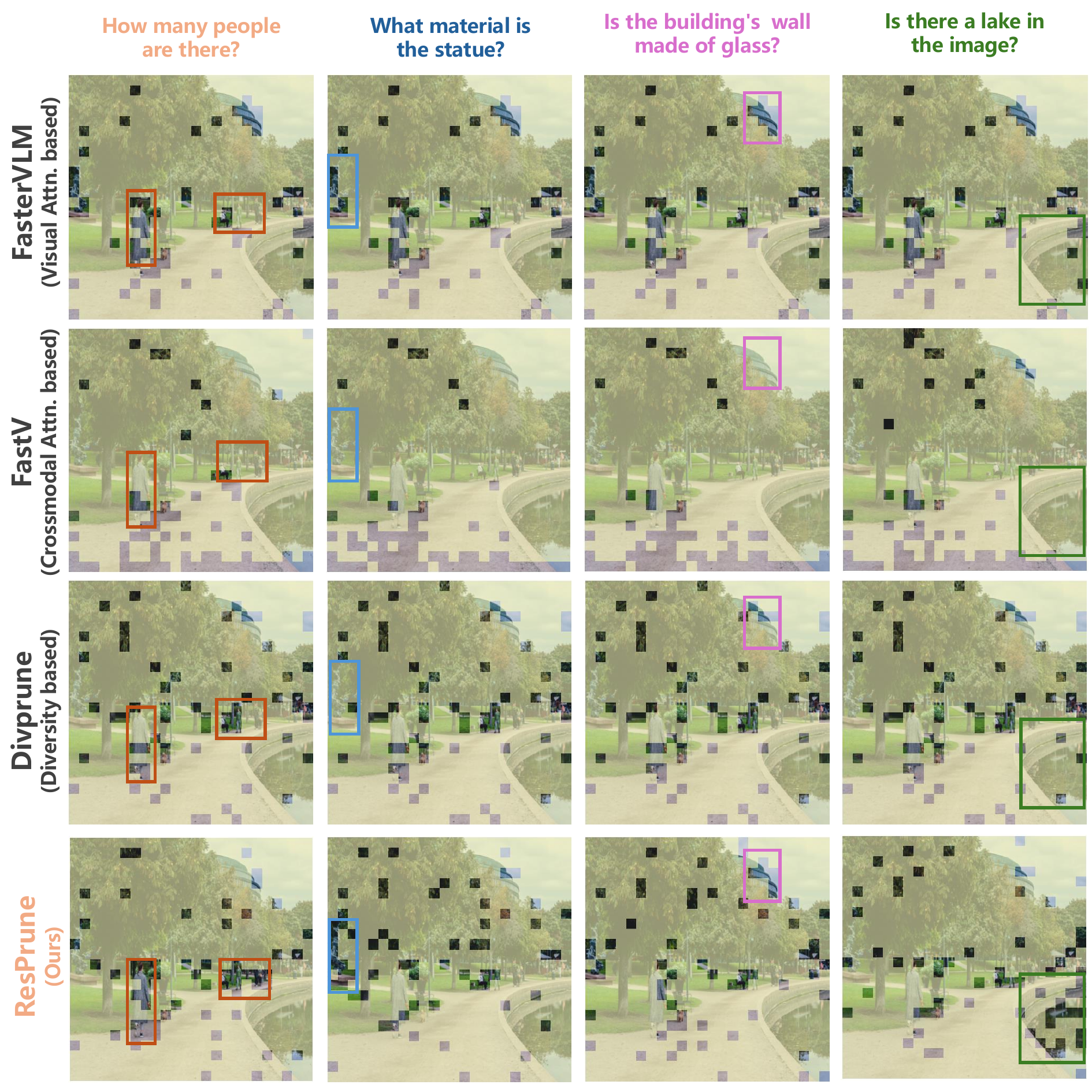}
    \caption{Comparison of visual token pruning behaviors across different methods under varying textual queries for a fixed image. Masked patches indicate pruned tokens, while unmasked patches represent the retained ones. Answer-relevant regions are highlighted with colored bounding boxes.}
  \label{fig:pruning_compare}
\end{figure}

While existing approaches have achieved encouraging empirical results, they each exhibit notable limitations.
Visual attention-based methods tend to preserve large clusters of tokens around visually salient regions, such as a dominant object, which often leads to semantic redundancy and insufficient contextual coverage. Moreover, as these methods rely solely on visual cues, they lack crossmodal guidance and may fail when visually salient regions do not align with the input instruction (see the 1st row in Fig.~\ref{fig:pruning_compare}).
Although crossmodal attention-based methods explicitly model text-image interactions, they also face several challenges.Prior studies~\cite{vispruner, dart} have shown that text-to-visual attention in shallow LLM layers often exhibits strong positional bias, favoring visual tokens that are closer to the textual queries in the input sequence (see the 2nd row in Fig.~\ref{fig:pruning_compare}).
In addition, these methods require all visual tokens to pass through multiple LLM layers, and extracting attention scores further complicates the use of optimized kernels such as FlashAttention~\cite{flashattention}, thereby diminishing the computational benefits of token pruning.
Diversity-based methods mitigate redundancy by encouraging broad coverage in the visual embedding space, but they are typically text-agnostic.
That is, for a fixed image, the pruning outcome remains static regardless of the input instruction, which can be suboptimal when a query requires fine-grained perception of specific objects or regions (see the 3rd row in Fig.~\ref{fig:pruning_compare}).

Motivated by the limitations of existing approaches, we revisit visual token pruning from a more principled perspective, focusing on three fundamental aspects.
First, we ask: \textbf{Should visual tokens be selected in a single pass based on token-wise importance scores?} One-shot selection strategies implicitly assume that a token's importance is independent of which other tokens are selected. However, in practice, the utility of a token is typically context-dependent: a token that seems informative in isolation may become redundant when similar tokens are already retained, whereas another token may gain importance by providing complementary information not yet captured. Therefore, token utility should be evaluated dynamically through a progressive selection process, rather than determined by a single-pass top-$K$ selection based on static importance scores.
This naturally leads to our second consideration: \textbf{How should the value of a candidate token be measured given an existing subset?} Prior work often approximates marginal utility using pairwise similarity measures. In contrast, we advocate a more explicit and principled criterion: a candidate token is valuable if it cannot be well represented by the current subset. In other words, effective pruning should directly quantify the residual information that remains unexplained by the selected tokens. 
Finally, we ask: \textbf{How can textual relevance be incorporated into token selection without entering the LLM?} In modern LVLMs, visual and textual tokens are projected into a shared embedding space before entering the language model. This design enables text–visual alignment to be estimated directly via similarity in the input space, allowing for instruction-aware pruning without incurring additional LLM inference costs.

Based on these considerations, we propose \textbf{ResPrune}, a novel training-free framework for visual token pruning in LVLMs. We formulate token pruning as a subspace reconstruction problem in the LLM's input embedding space. To address this problem, we introduce a greedy approximation strategy that iteratively selects the visual token with the highest residual energy with respect to the subspace spanned by the currently selected tokens. To efficiently incorporate crossmodal relevance, we further apply a text-guided gating mechanism that modulates the residual energy scores without requiring any attention weights from the LLM. By jointly considering visual coverage and instruction relevance, ResPrune progressively expands a compact token subset that preserves both global visual context and task-critical visual evidence. 
Extensive experiments across diverse LVLM backbones, downstream tasks, and token pruning ratios demonstrate that ResPrune consistently outperforms prior pruning strategies, achieving strong and robust performance. For example, on LLaVA-1.5-7B, ResPrune preserves 99.3\% of the original performance while pruning 77.8\% of visual tokens.

\noindent In summary, our main contributions are threefold:
\begin{itemize}
\item We propose ResPrune, a training-free visual token pruning framework for LVLMs that can be seamlessly integrated into existing pipelines without any model modification or retraining.

\item We formulate token selection as a text-aware subspace reconstruction problem and introduce a greedy selection algorithm based on residual energy, offering a principled and dynamic alternative to attention-based one-shot and similarity-based iterative pruning methods.

\item We conduct extensive evaluations across diverse LVLM backbones (LLaVA-1.5~\cite{llava1.5}, LLaVA-NeXT~\cite{llavanext}, and Qwen2.5-VL~\cite{qwen2.5}) and downstream benchmarks, demonstrating consistently superior trade-offs between performance retention and token reduction.
\end{itemize}

\section{Related Work}

\subsection{The Development of LVLMs}
The remarkable success of Large Language Models (LLMs) \cite{gpt1, gpt2, gpt3} has led to a growing trend of extending their powerful reasoning capabilities to the visual modality, giving rise to Large Vision-Language Models (LVLMs) \cite{llava, llava1.5, llavanext, instructblip}. A typical LVLM architecture consists of three main components: a vision encoder, a projector, and an LLM. The vision encoder first extracts visual representations from an input image, which are then mapped into the LLM’s input embedding space through a projector. These visual tokens are concatenated with textual tokens along the sequence dimension and jointly processed by the LLM to enable multimodal understanding and text generation. Despite their strong performance across a wide range of multimodal tasks, LVLMs typically produce a large number of visual tokens for an input image. For example, LLaVA-1.5 encodes a single image into 576 visual tokens \cite{llava1.5}. As LVLMs evolve toward finer-grained visual understanding, recent models have adopted increasingly sophisticated encoding strategies that further inflate the number of visual tokens. For example, LLaVA-NeXT \cite{llavanext} employs sub-image partitioning schemes, producing up to 2,880 visual tokens for a single image, while Qwen2.5-VL \cite{qwen2.5} adopts dynamic resolution strategies that can generate over 10,000 tokens under high-resolution settings.

Since all visual tokens are fed into the LLM for joint processing, and the computational complexity of self-attention scales quadratically with the input sequence length, the rapidly growing number of visual tokens leads to substantial inference overhead.
This significantly hinders the applicability of LVLMs in latency-sensitive scenarios such as robotics control \cite{robotics} and autonomous driving \cite{autonomous_drive}.

\subsection{Visual Token Pruning for LVLMs}
To mitigate the computational burden caused by the large number of visual tokens, a variety of studies have explored visual token pruning in LVLMs.

The first line of work leverages visual attention from the vision encoder to identify informative tokens \cite{prumerge, visionzip, vispruner}. For example, FasterVLM \cite{fastervlm} retains visually salient tokens based on CLS-to-patch attention scores. LLaVA-Prumerge \cite{prumerge} extends this idea by not only selecting tokens via CLS attention but also merging the discarded tokens into the retained ones to reduce information loss. 
The second line of work performs visual token pruning based on crossmodal attention within the LLM \cite{fastv, sparsevlm, pdrop}. For instance, FastV \cite{fastv} selects visual tokens according to the attention weights they receive from textual tokens. SparseVLM \cite{sparsevlm} further refines this approach by first identifying textual tokens that are strongly grounded in the image and then focusing on their corresponding crossmodal attention patterns to guide token selection. 
The third line of work explicitly models redundancy among visual tokens to encourage diversity in the retained token subset \cite{divprune, dart, scope}. For example, DivPrune \cite{divprune} iteratively selects tokens that exhibit the lowest maximum similarity to those already selected, thereby avoiding over-representation of similar content. In contrast, DART \cite{dart} first identifies a small set of anchor tokens and then completes the subset by minimizing the maximum similarity to these anchors in a single pass.

Although these methods have demonstrated promising results, they each have limitations. 
Attention-based methods often rely on one-shot importance scores, which tend to preserve clusters of highly correlated tokens, leading to semantic redundancy.
On the other hand, diversity-based methods emphasize geometric dispersion in the embedding space but typically overlook task semantics, making them less effective at retaining instruction-relevant content. These limitations highlight the need for more principled token pruning strategies that can jointly address redundancy, visual coverage, and instruction relevance in a unified framework. 

\section{Method}

In this section, we present \textbf{ResPrune}, a novel visual token pruning framework for efficient LVLM inference.
We begin by formulating visual token pruning as a constrained optimization problem (Section~\ref{sec:problem_formulation}).
We then derive a tractable surrogate objective from a subspace reconstruction perspective and introduce a greedy selection algorithm based on residual energy (Sections~\ref{sec:subspace_surrogate} and~\ref{sec:residual_energy}).
Next, we describe how textual guidance is incorporated to enable instruction-aware token selection (Section~\ref{sec:text_guided_residual}).
Finally, we detail the practical implementation of ResPrune and analyze its computational efficiency (Sections~\ref{sec:algorithm} and~\ref{sec:complexity}).

\subsection{Problem Formulation}
\label{sec:problem_formulation}

A typical LVLM consists of three components: a vision encoder, a vision-language projector, and an LLM.
Given an input image, the vision encoder extracts dense visual features, which are then mapped by the projector into the LLM's input embedding space.
We denote the resulting visual token sequence as
\begin{equation}
\mathbf{V} = [\mathbf{v}_1,\ldots,\mathbf{v}_T] \in \mathbb{R}^{T \times d},
\end{equation}
where $T$ is the number of visual tokens and $d$ is the embedding dimension of the LLM.
Given an input instruction or question, we denote the embedded textual token sequence as
\begin{equation}
\mathbf{U} = [\mathbf{u}_1,\ldots,\mathbf{u}_L] \in \mathbb{R}^{L \times d},
\end{equation}
where $L$ is the number of textual tokens.
Let $\mathcal{M}_{\theta}$ denote the LLM as a conditional generation model parameterized by $\theta$.
When provided with the full visual and textual inputs, the model generates an output
\begin{equation}
\mathbf{y} = \mathcal{M}_{\theta}(\mathbf{V}, \mathbf{U}),
\end{equation}
which represents the textual response conditioned on both the visual content and textual instruction.

The goal of visual token pruning is to select a subset of $k$ visual tokens, with $k \ll T$, that can replace the full visual token sequence while preserving task performance.
Formally, we seek an index set $S \subseteq \{1,\ldots,T\}$ with $|S|=k$, and form the pruned visual token sequence
\begin{equation}
\mathbf{V}_S = [\mathbf{v}_i]_{i \in S} \in \mathbb{R}^{k \times d},
\end{equation}
which replaces $\mathbf{V}$ as the visual input to the LLM:
\begin{equation}
\hat{\mathbf{y}} = \mathcal{M}_{\theta}(\mathbf{V}_S, \mathbf{U}).
\end{equation}

Ideally, visual token pruning should preserve the model behavior, such that $\hat{\mathbf{y}}$ remains close to $\mathbf{y}$ under an appropriate task loss.
This objective can be formulated as the following constrained optimization problem:
\begin{equation}
\min_{S} \ 
\mathcal{L}\!\left(\mathcal{M}_{\theta}(\mathbf{V}_S,\mathbf{U}), \ \mathcal{M}_{\theta}(\mathbf{V},\mathbf{U})\right)
\quad \text{s.t.} \quad |S| = k,
\label{eq:ideal_obj}
\end{equation}
where $\mathcal{L}$ measures the discrepancy between the model outputs with and without visual token pruning. Directly optimizing Eq.~\eqref{eq:ideal_obj} is intractable in practice, as it would require repeatedly querying the full LLM with different token subsets and backpropagating through the entire model.

\subsection{Surrogate Objective: Subspace Reconstruction}
\label{sec:subspace_surrogate}

To design a tractable surrogate for Eq.~\eqref{eq:ideal_obj}, we reformulate visual token pruning as a subspace reconstruction problem in the LLM's input embedding space. In other words, we aim to identify a visual token subset whose span can accurately approximate the full token set, as measured by the reconstruction error in the embedding space. 

Given a candidate index set $S \subseteq \{1,\ldots,T\}$, we define the subspace spanned by the corresponding visual tokens as
\begin{equation}
\mathcal{W}(S) = \mathrm{Span}\{\mathbf{v}_i : i \in S\} \subseteq \mathbb{R}^{d}.
\end{equation}
This subspace captures the set of directions that can be expressed as linear combinations of these tokens.
If the remaining tokens lie close to $\mathcal{W}(S)$, then their representations can be well approximated by the selected tokens, indicating that the overall visual information is largely preserved by $\mathbf{V}_S$.

Let $P_S$ denote the orthogonal projection operator onto $\mathcal{W}(S)$.
The quality of a candidate subset $S$ can be evaluated by the total reconstruction error of the full token set with respect to the subspace $\mathcal{W}(S)$:
\begin{equation}
\big\| \mathbf{V} - P_S \mathbf{V} \big\|_F^2,
\end{equation}
where $\|\cdot\|_F$ denotes the Frobenius norm.
Based on this formulation, we define the following surrogate objective for visual token pruning:
\begin{equation}
\min_{S} \ \big\| \mathbf{V} - P_S \mathbf{V} \big\|_F^2
\quad \text{s.t.} \quad |S| = k.
\label{eq:subspace_obj}
\end{equation}
Intuitively, this objective seeks a subset of $k$ visual tokens whose span best reconstructs the original token set (or equivalently, the subset that maximizes the explained variance of the full visual representations). Unlike similarity-based selection, this formulation directly captures the global structure of the embedding space, rather than relying on pairwise distances.

\subsection{Greedy Subspace Expansion via Residual Energy}
\label{sec:residual_energy}
The surrogate objective in Eq.~\eqref{eq:subspace_obj} defines a combinatorial optimization problem that is hard to solve exactly.
We therefore adopt a greedy approximation strategy that progressively expands the selected token subset based on residual energy.

Let $S$ denote the current set of selected token indices, and let $\mathcal{W}(S)$ be the subspace spanned by the corresponding visual tokens.
For any candidate visual token $\mathbf{v} \in \mathbb{R}^{d}$, we decompose it into two orthogonal components:
\begin{equation}
\mathbf{v}
=
P_S \mathbf{v}
+
\mathbf{r}_S(\mathbf{v}),
\end{equation}
where $P_S \mathbf{v}$ is the orthogonal projection of $\mathbf{v}$ onto $\mathcal{W}(S)$, and $\mathbf{r}_S(\mathbf{v}) = \mathbf{v} - P_S \mathbf{v}$ is the residual component orthogonal to the current subspace.
The residual captures the portion of $\mathbf{v}$ that cannot be explained by the tokens selected so far.

We define the residual energy of token $\mathbf{v}$ as:
\begin{equation}
E_S(\mathbf{v}) = \big\| \mathbf{r}_S(\mathbf{v}) \big\|_2^2 = \big\| \mathbf{v} - P_S \mathbf{v} \big\|_2^2,
\label{eq:residual_energy}
\end{equation}
which quantifies the amount of novel information that $\mathbf{v}$ contributes beyond the span of the current subset. Tokens with higher residual energy indicate directions in the embedding space that are underrepresented by $\mathcal{W}(S)$. 

Starting from an initial seed token, we iteratively select the next token with the highest residual energy :
\begin{equation}
i^\star = \arg\max_{i \notin S} \ E_S(\mathbf{v}_i).
\end{equation}
This greedy rule provides a tractable approximation to the objective in Eq.~\eqref{eq:subspace_obj} by prioritizing tokens that contribute the largest unexplained components with respect to the current subspace, thereby progressively improving the reconstruction capability of $\mathcal{W}(S)$.

In practice, residual energies can be computed efficiently by maintaining an orthonormal basis for the evolving subspace $\mathcal{W}(S)$, as detailed in Section~\ref{sec:algorithm}.

\subsection{Conditioning on Textual Relevance}
\label{sec:text_guided_residual}

The greedy selection strategy described in Section~\ref{sec:residual_energy} is task-agnostic.
However, in practice, the importance of visual tokens is inherently conditioned on the input instruction.
Different queries may emphasize different aspects of the same image, making it desirable to bias token selection toward instruction-relevant visual content.

We exploit the fact that visual and textual tokens reside in a shared space (i.e., the LLM's input embedding space), which enables direct comparison between modalities.
Given a visual token $\mathbf{v}_i$ and a textual token set $\mathbf{U} = [\mathbf{u}_1,\ldots,\mathbf{u}_L]$, we define the textual relevance score of $\mathbf{v}_i$ as:
\begin{equation}
R(\mathbf{v}_i, \mathbf{U})
=
\max_{1 \le j \le L}
\ \max(0,\, \mathrm{sim}(\mathbf{v}_i, \mathbf{u}_j)),
\label{eq:text_relevance}
\end{equation}
where $\mathrm{sim}(\cdot,\cdot)$ denotes cosine similarity.

To incorporate textual relevance into the greedy selection process, we modulate the residual energy of each candidate token by a non-negative relevance weight.
Specifically, we define the text-conditioned residual energy as
\begin{equation}
\widetilde{E}_S(\mathbf{v}_i)
=
E_S(\mathbf{v}_i) \cdot g\!\left(R(\mathbf{v}_i, \mathbf{U})\right),
\label{eq:text_guided_energy}
\end{equation}
where $E_S(\mathbf{v}_i)$ is the residual energy defined in Eq.~\eqref{eq:residual_energy} and $g(\cdot)$ is a monotonically increasing gating function:
\begin{equation}
g(r) = r^{\alpha}, \quad \alpha \ge 0,
\end{equation}
where $\alpha$ controls the strength of textual guidance.
The greedy selection rule is then modified to
\begin{equation}
i^\star = \arg\max_{i \notin S} \ \widetilde{E}_S(\mathbf{v}_i),
\end{equation}
which prioritizes tokens that not only contribute novel directions to the subspace but also exhibit strong alignment with the textual instruction. This design enables instruction-aware token selection without relying on any attention weights from the LLM, and therefore remains fully compatible with FlashAttention.

\begin{algorithm}[t]
\caption{ResPrune (Practical Implementation)}
\label{alg:resprune_impl}
\begin{algorithmic}[1]
\STATE \textbf{Input:} visual tokens $\mathbf{V}=[\mathbf{v}_1,\ldots,\mathbf{v}_T]\in\mathbb{R}^{T\times d}$, text prompt $\mathbf{x}$, token budget $k$, guidance strength $\alpha\ge0$
\STATE \textbf{Output:} selected indices $S\subseteq\{1,\ldots,T\}$ with $|S|=k$ 

\vspace{0.25em}
\STATE \textbf{// Seed token initialization}
\STATE Choose an initial seed index $i_0$ using any valid initialization strategy.
\STATE $S \leftarrow \{i_0\}$

\vspace{0.25em}
\STATE \textbf{// Text cleaning and embedding}
\STATE $\tilde{\mathbf{x}} \leftarrow \mathrm{CleanText}(\mathbf{x})$ \hfill // extract nouns
\STATE $\mathbf{U} \leftarrow \mathrm{EmbedText}(\tilde{\mathbf{x}})$ \hfill // textual token embeddings

\vspace{0.25em}
\STATE \textbf{// Textual relevance computation}
\STATE $r_i \leftarrow \max_{j}\max\!\bigl(0,\mathrm{cos}(\mathbf{v}_i,\mathbf{u}_j)\bigr),\ \forall i,$ if $\mathbf{U}\neq\emptyset$; otherwise $r_i \leftarrow 1,\ \forall i$
\STATE $w_i \leftarrow (r_i+\varepsilon)^{\alpha},\ \forall i$ \hfill // $\varepsilon>0$ for stability

\vspace{0.25em}
\STATE \textbf{// Greedy subspace expansion}
\STATE Initialize an orthonormal basis $\mathbf{Q}$ from $\mathbf{v}_{i_0}$.
\STATE Initialize residual energies $e_i \leftarrow \|\mathbf{v}_i\|_2^2 - \|\mathbf{Q}^\top\mathbf{v}_i\|_2^2,\ \forall i$.
\WHILE{$|S|<k$}
    \STATE $i^\star \leftarrow \arg\max_{i\notin S} \ (e_i \cdot w_i)$
    \STATE $S \leftarrow S \cup \{i^\star\}$
    \STATE Update $\mathbf{Q}$ via Gram--Schmidt orthogonalization
    \STATE Update residual energies $\{e_i\}$ using the updated $\mathbf{Q}$
\ENDWHILE

\STATE \textbf{return} $S$
\end{algorithmic}
\end{algorithm}

\subsection{Practical Implementation}
\label{sec:algorithm}

Algorithm~\ref{alg:resprune_impl} outlines the practical implementation of ResPrune.
The algorithm follows the theoretical formulation in the previous sections and is designed to be training-free, efficient, and easily integrated into existing LVLM pipelines.
Below we highlight several key implementation aspects.

\paragraph{Seed token initialization.}
The greedy selection procedure requires an initial seed token to bootstrap subspace construction.
ResPrune does not impose any restriction on how this seed token is chosen.
In practice, various strategies can be adopted, such as attention-based heuristics or feature norm–based selection. The specific strategy used in our experiments is specified in Section~\ref{sec:implementation_details}.

\paragraph{Text preprocessing and relevance computation.}
Prior to computing textual relevance scores, we apply lightweight preprocessing to the input prompt. Specifically, we remove answer-formatting instructions (e.g., “answer the question using a single word or phrase”) and extract only noun tokens, which are more likely to correspond to visually grounded entities. If no valid text are found, ResPrune falls back to a task-agnostic variant by assigning uniform relevance weights across all visual tokens.

\paragraph{Efficient residual energy updates.}
A key advantage of ResPrune is that residual energies can be updated efficiently without explicitly computing projection operators.
The algorithm maintains an orthonormal basis of the subspace spanned by the selected tokens and updates residual energies incrementally using inner products.
At each iteration, only a single Gram--Schmidt orthogonalization step is required to incorporate the newly selected token, followed by a lightweight update of residual energies for the remaining candidates.

All operations of ResPrune are performed prior to LLM inference, enabling seamless integration with optimized attention kernels and existing LVLM architectures.

\subsection{Theoretical Efficiency Analysis}
\label{sec:complexity}

We analyze the computational efficiency of ResPrune from two complementary perspectives:
(1) the cost of the pruning procedure itself, and
(2) the inference savings achieved by reducing the number of visual tokens.

\paragraph{Complexity of ResPrune.}
ResPrune consists of two stages: (i) computing textual relevance weights and (ii) greedy subspace expansion.
For textual relevance, we compute similarities between all visual tokens and the extracted text tokens, which can be implemented as a single matrix multiplication followed by a max operation. This incurs a time complexity of $\mathcal{O}(TLd)$.
For greedy selection, ResPrune performs $k$ iterations. At each iteration, the residual-energy update requires one inner-product pass between the newly added basis direction and all visual tokens, resulting in $\mathcal{O}(Td)$ per iteration and $\mathcal{O}(kTd)$ in total. 
Combining the above, the overall time complexity of ResPrune is
\begin{equation}
\mathcal{O}(TLd) + \mathcal{O}(kTd + k^2d).
\end{equation}
In the common regime $T\gg k$, it simplifies to $\mathcal{O}(TLd + kTd)$.

\paragraph{Efficiency gains from visual token reduction.}
In LVLMs, the LLM constitutes the dominant source of computational overhead during inference.
Specifically, during the prefill stage, all visual and textual tokens are processed in a single forward pass to cache intermediate states for subsequent decoding.
Since every token participates in self-attention, the computational cost of this stage dominates the overall inference complexity.

Let $T$ and $L$ denote the numbers of visual and textual tokens, respectively.
The per-layer FLOPs of the LLM in the prefill stage can be expressed as
\begin{equation}
8(T+L)d^2 + 4(T+L)^2d + 6(T+L)dm,
\end{equation}
where $m$ is the intermediate dimension of its MLP layers.
In typical settings, the number of visual tokens substantially exceeds that of textual tokens ($T \gg L$), causing the quadratic self-attention term to be dominated by visual tokens.

By applying ResPrune, the number of visual tokens processed by the LLM is reduced from $T$ to $k$.
Accordingly, the per-layer FLOPs in the prefill stage become
\begin{equation}
8(k+L)d^2 + 4(k+L)^2d + 6(k+L)dm.
\end{equation}
When $k \ll T$ and $T \gg L$, this reduction significantly lowers the dominant quadratic term in self-attention, resulting in substantial computational savings during LLM inference.

\paragraph{Overall efficiency.}
Although ResPrune introduces additional preprocessing steps, its cost is outweighed by the substantial reduction in the quadratic complexity of LLM's self-attention. As a result, ResPrune effectively shifts the computational burden from the LLM to a lightweight token selection stage, achieving improved inference efficiency.

\section{Experiments}

\begin{table*}[ht]
\centering
\caption{Performance comparison on LLaVA-1.5-7B under different visual token pruning ratios.
``Rel. Perf.'' denotes the average relative performance with respect to the full-token baseline where higher values indicate better performance retention. Best results under each pruning ratio are highlighted in bold.}
\label{tab:case_struct}
\resizebox{\textwidth}{!}{%
\begin{tabular}{l|c|ccccccccc}
\toprule
\rowcolor[HTML]{F2F3F5} 
\textbf{Method}      & \textbf{Rel. Perf.} & \textbf{MME}  & \textbf{GQA}  & \textbf{SQA-I} & \textbf{POPE} & \textbf{TextVQA} & \textbf{VizWiz} & \textbf{VQA-v2} & \textbf{MMB-en} & \textbf{MM-Vet} \\
\midrule
\midrule
LLaVA-1.5-7B \cite{llava1.5}         & 100\%              & 1862          & 61.9          & 69.5           & 85.9          & 58.2             & 50.0            & 78.4            & 64.7            & 31.3            \\
\midrule
\rowcolor[HTML]{F5F6F7}
\multicolumn{11}{c}{\cellcolor[HTML]{F5F6F7}Retain 192 visual tokens (pruning ratio = 66.7\%)}       \\
\midrule
ToMe (ICLR2023) \cite{tome}      & 89.9                & 1563          & 54.3          & 65.2           & 72.4          & 52.1             & 50.0            & 68.0            & 60.5            & 26.8            \\
FastV (ECCV2024) \cite{fastv}     & 89.5                & 1612          & 52.7          & 67.3           & 64.8          & 52.5             & 50.8            & 67.1            & 61.2            & 27.7            \\
PruMerge (ICCV2025) \cite{prumerge}  & 91.5                & 1632          & 54.3          & 67.9           & 71.3          & 54.3             & 50.1            & 70.6            & 59.6            & -               \\
SparseVLM (ICML2025) \cite{sparsevlm} & 96.6                & 1721          & 57.6          & 69.1           & 83.6          & 56.1             & 50.5            & 75.6            & 62.5            & 31.5            \\
HiRED (AAAI2025) \cite{hired}   & 94.6                & 1737          & 58.7          & 68.4           & 82.8          & 47.4             & 50.1            & 74.9            & 62.8            & -               \\
DART (EMNLP2025) \cite{dart}   & 98.6                & \textbf{1856} & 58.9          & \textbf{69.8}  & 82.8          & 57.4             & 51.1            & 76.7            & 63.6            & 31.5            \\
VisionZIP (CVPR2025) \cite{visionzip} & 98.2                & 1767          & 59.3          & 68.9           & 85.3          & 57.8             & 50.9            & 76.5            & 63.0            & 31.7            \\
DivPrune (CVPR2025) \cite{divprune}  & 97.9                & 1751          & 59.8          & 69.2           & 87.1          & 54.8             & 51.2            & 76.8            & 63.0            & 32.0            \\
PDrop (CVPR2025) \cite{pdrop}     & 96.7                & 1766          & 57.1          & 68.8           & 82.3          & 56.1             & 51.1            & 75.1            & 63.2            & 30.5            \\
SCOPE (NIPS2025) \cite{scope}     & 98.9                & 1804          & 60.1          & 68.8           & 86.4          & 57.7             & 51.0            & 76.9            & 63.6            & 32.5            \\
\rowcolor[HTML]{F0F4FF} 
ResPrune (Ours)      & \textbf{99.4}       & 1795          & \textbf{60.6} & 69.1           & \textbf{87.6} & \textbf{58.1}    & \textbf{51.2}   & \textbf{77.4}   & \textbf{63.8}   & \textbf{32.8}   \\
\midrule
\rowcolor[HTML]{F5F6F7}
\multicolumn{11}{c}{\cellcolor[HTML]{F5F6F7}Retain 128 visual tokens (pruning ratio = 77.8\%)}                                                                                       \\
\midrule
ToMe (ICLR2023) \cite{tome}     & 82.9                & 1343          & 52.4          & 59.6           & 62.8          & 49.1             & 50.1            & 63.0            & 53.3            & 27.2            \\
FastV (ECCV2024) \cite{fastv}     & 83.9                & 1490          & 49.6          & 60.2           & 59.6          & 50.6             & 51.3            & 61.8            & 56.1            & 28.1            \\
PruMerge (ICCV2025) \cite{prumerge}  & 89.5                & 1554          & 53.3          & 67.1           & 67.2          & 54.3             & 50.3            & 68.8            & 58.1            & -               \\
TRIM (ICCL2025) \cite{trim}     & 96.5                & 1743          & 58.4          & 68.6           & 85.3          & 52.2             & 51.6            & 75.4            & 63.0            & 29.9            \\
VisPruner (ICCV2025) \cite{vispruner} & 97.9                & 1762          & 58.2          & 69.1           & 84.6          & 57.0             & 52.7            & 75.8            & 62.7            & \textbf{33.7}   \\
SparseVLM (ICML2025) \cite{sparsevlm} & 94.5                & 1696          & 56.0          & 67.1           & 80.5          & 54.9             & 51.4            & 73.8            & 60.0            & 29.0            \\
HiRED (AAAI2025) \cite{hired}     & 93.2                & 1710          & 57.2          & 68.1           & 79.8          & 46.1             & 51.3            & 73.4            & 61.5            & -               \\
DART (EMNLP2025) \cite{dart}    & 97.0                & \textbf{1845} & 57.9          & 69.1           & 80.1          & 56.4             & 51.7            & 75.9            & 60.7            & 30.9            \\
VisionZIP (CVPR2025) \cite{visionzip} & 97.1                & 1762          & 57.6          & 68.7           & 83.3          & 56.9             & 51.6            & 75.6            & 62.1            & 32.6            \\
DivPrune (CVPR2025) \cite{divprune}  & 97.2                & 1694          & 59.4          & 68.5           & 87.0          & 54.5             & \textbf{52.7}   & 76.0            & 61.5            & 30.7            \\
PDrop (CVPR2025) \cite{pdrop}    & 94.6                & 1644          & 56.0          & 68.3           & 82.3          & 55.1             & 51.0            & 72.9            & 61.1            & 30.8            \\
SCOPE (NIPS2025) \cite{scope}     & 98.3                & 1776          & 59.7          & 68.4           & 86.1          & 57.2             & 52.0            & 76.1            & 62.5            & 31.4            \\
AdaPrune (NIPS2025) \cite{adaprune}  & 98.3                & 1755          & 59.3          & 68.5           & 86.5          & 57.0             & 52.6            & 76.4            & 62.3            & -               \\
\rowcolor[HTML]{F0F4FF} 
ResPrune (Ours)      & \textbf{99.3}       & 1792          & \textbf{60.1} & \textbf{69.3}  & \textbf{87.6} & \textbf{57.8}    & 52.5            & \textbf{76.5}   & \textbf{63.0}   & 31.8            \\
\midrule
\rowcolor[HTML]{F5F6F7}
\multicolumn{11}{c}{\cellcolor[HTML]{F5F6F7}Retain 64 visual tokens (pruning ratio = 88.9\%)}   \\
\midrule
ToMe (ICLR2023) \cite{tome}      & 73.8                & 1138          & 48.6          & 50.0           & 52.5          & 45.3             & 49.8            & 57.1            & 43.7            & 25.8            \\
FastV (ECCV2024) \cite{fastv}     & 74.9                & 1256          & 46.1          & 51.1           & 48.0          & 47.8             & 50.8            & 55.0            & 48.0            & 26.7            \\
PruMerge (ICCV2025) \cite{prumerge}  & 88.2                & 1549          & 51.9          & 68.1           & 65.3          & 54.0             & 50.1            & 67.4            & 55.3            & -               \\
TRIM (ICCL2025) \cite{trim}      & 94.4                & 1680          & 56.6          & 69.0           & 85.9          & 49.7             & 51.1            & 72.4            & 60.9            & 24.8            \\
VisPruner (ICCV2025) \cite{vispruner} & 95.3                & 1674          & 55.4          & 69.1           & 80.4          & 55.8             & 53.3            & 72.7            & 61.3            & 31.7            \\
SparseVLM (ICML2025) \cite{sparsevlm} & 88.2                & 1505          & 52.7          & 62.2           & 75.1          & 51.8             & 50.1            & 68.2            & 56.2            & 24.9            \\
HiRED (AAAI2025) \cite{hired}     & 89.5                & 1599          & 54.6          & 68.2           & 73.6          & 44.2             & 50.2            & 69.7            & 60.2            & -               \\
DART (EMNLP2025) \cite{dart}    & 94.3                & \textbf{1765} & 55.9          & \textbf{69.8}  & 73.9          & 54.4             & 51.6            & 72.4            & 60.6            & 26.5            \\
VisionZIP (CVPR2025) \cite{visionzip} & 94.4                & 1690          & 55.1          & 69.0           & 77.0          & 55.5             & 52.9            & 72.4            & 60.1            & \textbf{31.7}   \\
DivPrune (CVPR2025) \cite{divprune} & 95.7                & 1657          & 57.9          & 67.9           & 85.6          & 52.9             & 53.6            & 74.1            & 60.2            & 29.4            \\
PDrop (CVPR2025) \cite{pdrop}     & 76.3                & 1092          & 41.9          & 68.6           & 55.9          & 45.9             & 50.7            & 69.2            & 33.3            & 30.7            \\
SCOPE (NIPS2025) \cite{scope}     & 96.8                & 1698          & 58.3          & 68.6           & 83.9          & 56.6             & 52.7            & 73.9            & \textbf{61.7}   & 30.4            \\
AdaPrune (NIPS2025) \cite{adaprune}  & 97.1                & 1715          & 57.4          & 68.8           & 84.8          & 56.0             & \textbf{53.9}   & 74.7            & 61.2            & -               \\
\rowcolor[HTML]{F0F4FF} 
ResPrune (Ours)      & \textbf{98.0}       & 1740          & \textbf{58.6} & 69.0           & \textbf{87.5} & \textbf{57.2}    & 53.6            & \textbf{74.9}   & 60.7            & 30.2    \\
\bottomrule
\end{tabular}
}
\end{table*}

\subsection{Experimental Settings}

\subsubsection{Implementation Details}
\label{sec:implementation_details}
We evaluate the proposed ResPrune on three representative LVLM backbones: \textbf{LLaVA-1.5-7B} \cite{llava1.5}, \textbf{LLaVA-NeXT-7B} \cite{llavanext}, and \textbf{Qwen2.5-VL-7B} \cite{qwen2.5}. These models cover different visual encoding strategies.
LLaVA-1.5 follows a traditional fixed-resolution design, encoding each input image into 576 visual tokens.
LLaVA-NeXT adopts an tile cropping strategy that partitions an image into multiple sub-images and encodes them independently, yielding up to 2880 visual tokens per image.
Qwen2.5-VL further generalizes this by enabling dynamic-resolution encoding via finetuning the vision encoder, producing over 10,000 visual tokens for high-resolution inputs.

For LLaVA-1.5 and LLaVA-NeXT, we set the textual guidance strength to $\alpha=0.75$, and select the initial seed token as the one receiving the highest CLS attention in the vision encoder.
For Qwen2.5-VL, which lacks a CLS token, we instead choose the token with the highest $\ell_2$ norm as the seed and reduce the textual guidance strength to $\alpha=0.3$ to accommodate the denser token distributions.

In the text preprocessing stage, we apply regular expressions to remove instruction formatting patterns (e.g., ``Answer using a single word or phrase'') from the prompt. The remaining text is then passed through a lightweight NLP parser (\texttt{en\_core\_web\_sm} \cite{en_core_web_sm}) to extract noun tokens.

\subsubsection{Evaluation Benchmarks} 
We evaluate ResPrune across nine widely adopted vision-language benchmarks spanning diverse task categories. These include VQAv2 \cite{vqav2}, VizWiz \cite{vizwiz}, and GQA \cite{gqa} for general visual question answering; SQA-I \cite{sqa} for multimodal knowledge reasoning; TextVQA \cite{textvqa} for scene text understanding; POPE \cite{pope} for evaluating object hallucination; MME \cite{mme}, MMB-en \cite{mmb}, and MM-Vet \cite{mmvet} for comprehensive assessment of LVLM capabilities.

\subsubsection{Comparison Methods}
We compare ResPrune with a range of recent visual token pruning approaches published at major AI conferences, including visual attention-based methods \cite{tome, prumerge, vispruner, hired, visionzip}, crossmodal attention-based approaches \cite{fastv, sparsevlm, trim, pdrop}, and diversity-based strategies \cite{dart, divprune, scope, adaprune}. Following standard evaluation protocols in prior work, all baselines are tested under multiple token budgets to ensure a fair and comprehensive comparison. 

\subsection{Main Results}

\begin{table*}[ht]
\centering
\caption{Performance comparison on LLaVA-NeXT-7B under different visual token pruning ratios.
``Rel. Perf.'' denotes the average relative performance with respect to the full-token baseline where higher values indicate better performance retention. Best results under each pruning ratio are highlighted in bold.}
\label{tab:next}
\resizebox{\textwidth}{!}{%
\begin{tabular}{l|c|cccccccc}
\toprule
\rowcolor[HTML]{F2F3F5} 
\textbf{Method}      & \textbf{Rel. Perf.} & \textbf{MME}  & \textbf{GQA}  & \textbf{SQA-I} & \textbf{POPE} & \textbf{TextVQA} & \textbf{VizWiz} & \textbf{VQA-v2} & \textbf{MMB-en} \\
\midrule
\midrule
LLaVA-NeXT-7B \cite{llavanext}        & 100\%               & 1842          & 64.3          & 70.2           & 86.5          & 61.3             & 55.2            & 81.3            & 67.9            \\
\midrule
\rowcolor[HTML]{F5F6F7}
\multicolumn{10}{c}{\cellcolor[HTML]{F5F6F7}Retain up to 640 visual tokens (pruning ratio  = 66.7\%)}                                                                      \\
\midrule
FastV (ECCV2024) \cite{fastv}     & 94.7                & 1807          & 58.9          & 67.4           & 79.5          & 58.1             & 53.9            & 77.0            & 63.1            \\
PruMerge (ICCV2025) \cite{prumerge}  & 96.6                & 1790          & 60.8          & 67.8           & 85.3          & 54.9             & 57.9            & 78.2            & 64.6            \\
PDrop (CVPR2025) \cite{pdrop}     & 95.7                & 1782          & 60.0          & 66.7           & 83.8          & 57.8             & 53.8            & 79.1            & 64.1            \\
SparseVLM (ICML2025) \cite{sparsevlm} & 96.9                & 1772          & 61.2          & 67.6           & 85.3          & 59.7             & 53.6            & 79.2            & 65.9            \\
DivPrune (CVPR2025) \cite{divprune}  & 97.2                & 1773          & 61.9          & 67.8           & 86.9          & 57.0             & 55.7            & 79.3            & 65.8            \\
VisionZIP (CVPR2025) \cite{visionzip} & 98.1                & 1782          & 61.3          & 68.1           & 86.2          & 59.9             & 57.1            & 79.1            & 66.3            \\
DART (EMNLP2025) \cite{dart}     & 97.5                & 1793          & 61.3          & 68.2           & 85.0          & 59.5             & 57.0            & 78.3            & 64.9            \\
\rowcolor[HTML]{F0F4FF} 
ResPrune (Ours)      & \textbf{99.6}       & \textbf{1821} & \textbf{63.3} & \textbf{69.5}  & \textbf{88.3} & \textbf{59.9}    & \textbf{57.9}   & \textbf{79.5}   & \textbf{66.6}   \\
\midrule
\rowcolor[HTML]{F5F6F7}
\multicolumn{10}{c}{\cellcolor[HTML]{F5F6F7}Retain up to 320 visual tokens (pruning ratio  = 88.9\%)}      \\
\midrule
FastV (ECCV2024) \cite{fastv}     & 80.7                & 1539          & 49.8          & 66.6           & 49.5          & 52.2             & 51.3            & 61.5            & 53.4            \\
PruMerge (ICCV2025) \cite{prumerge}  & 94.1                & 1744          & 58.8          & 68.1           & 79.5          & 54.0             & \textbf{57.7}   & 75.3            & 63.0            \\
PDrop (CVPR2025) \cite{pdrop}     & 83.5                & 1672          & 50.4          & 66.7           & 60.8          & 49.0             & 49.7            & 66.8            & 55.5            \\
SparseVLM (ICML2025) \cite{sparsevlm} & 93.1                & 1747          & 57.9          & 67.2           & 76.9          & 56.5             & 54.2            & 74.6            & 63.1            \\
DivPrune (CVPR2025) \cite{divprune}  & 96.6                & 1731          & 61.1          & 67.7           & 84.7          & 56.2             & 55.6            & 77.2            & 63.9            \\
VisionZIP (CVPR2025) \cite{visionzip} & 95.0                & 1698          & 59.3          & 67.3           & 82.1          & 58.9             & 56.2            & 76.2            & 63.1            \\
DART (EMNLP2025) \cite{dart}     & 94.8                & 1710          & 59.5          & 67.5           & 81.0          & 57.6             & 56.1            & 75.7            & 64.2            \\
\rowcolor[HTML]{F0F4FF} 
ResPrune (Ours)      & \textbf{98.1}       & \textbf{1780} & \textbf{62.4} & \textbf{69.2}  & \textbf{88.0} & \textbf{59.5}    & 56.6            & \textbf{77.9}   & \textbf{64.9}  \\
\bottomrule
\end{tabular}
}
\end{table*}

\begin{table}[ht]
\centering
\caption{Performance comparison on Qwen2.5-VL-7B under different visual token pruning ratios.
``Rel. Perf.'' denotes the average relative performance with respect to the full-token baseline where higher values indicate better performance retention. Best results under each pruning ratio are highlighted in bold.}
\label{tab:qwen}
\resizebox{\columnwidth}{!}{%
\begin{tabular}{l|c|ccccc}
\toprule
\rowcolor[HTML]{F2F3F5} 
\textbf{Method}     & \textbf{Rel. Perf.} & \textbf{MME}  & \textbf{POPE} & \textbf{TextVQA} & \textbf{MMB-en} & \multicolumn{1}{l}{\cellcolor[HTML]{F2F3F5}\textbf{GQA}} \\
\midrule
\midrule
Qwen2.5-VL-7B       & 100\%               & 2304          & 86.1          & 84.8             & 82.8            & 62.2                                                     \\
\midrule
\rowcolor[HTML]{F5F6F7}
\multicolumn{7}{c}{\cellcolor[HTML]{F5F6F7}Pruning ratio  = 66.7\%}                                                                                       \\
\midrule
FastV (ECCV2024)    & 92.3                & 2072          & 82.2          & 77.9             & 75.7            & 58.0                                                     \\
DivPrune (CVPR2025) & 96.7                & 2198          & 85.6          & 80.1             & \textbf{82.6}   & 59.0                                                     \\
\rowcolor[HTML]{F0F4FF} 
ResPrune (Ours)     & \textbf{98.4}       & \textbf{2320} & \textbf{87.3} & \textbf{80.3}    & 82.5            & \textbf{59.9}                                            \\
\midrule
\rowcolor[HTML]{F5F6F7} 
\multicolumn{7}{c}{\cellcolor[HTML]{F5F6F7}Pruning ratio  = 77.8\%}                                                                                       \\
\midrule
FastV (ECCV2024)    & 88.9                & 2036          & 80.7          & 69.0             & 74.9            & 56.7                                                     \\
DivPrune (CVPR2025) & 95.1                & 2153          & 85.5          & 76.6             & 81.6            & 58.6                                                     \\
\rowcolor[HTML]{F0F4FF} 
ResPrune (Ours)     & \textbf{96.9}       & \textbf{2246} & \textbf{87.3} & \textbf{78.0}    & \textbf{81.7}   & \textbf{59.4}                                            \\
\midrule
\rowcolor[HTML]{F5F6F7} 
\multicolumn{7}{c}{\cellcolor[HTML]{F5F6F7}Pruning ratio = 88.9\%}                                                                                        \\
\midrule
FastV (ECCV2024)    & 82.6                & 1940          & 78.6          & 60.3             & 69.2            & 51.9                                                     \\
DivPrune (CVPR2025) & 90.4                & 2051          & 83.7          & 67.2             & 79.4            & 56.9                                                     \\
\rowcolor[HTML]{F0F4FF} 
ResPrune (Ours)     & \textbf{92.2}       & \textbf{2127} & \textbf{85.2} & \textbf{69.0}    & \textbf{79.5}   & \textbf{57.7}   \\                  
\bottomrule
\end{tabular}
}
\end{table}

\subsubsection{With LLaVA-1.5-7B} 
We evaluate ResPrune on LLaVA-1.5-7B under three pruning ratios of 66.7\%, 77.8\%, and 88.9\%.
As shown in Table~\ref{tab:case_struct}, ResPrune consistently outperforms existing baselines across a wide range of benchmarks and pruning ratios.
At a moderate pruning ratio of 66.7\%, ResPrune achieves the highest overall performance retention (99.4\%), surpassing all competing methods.
Moreover, it yields the best results on a broad set of benchmarks, including GQA, POPE, TextVQA, VizWiz, VQA-v2, MMB-en, and MM-Vet.
Under a more aggressive pruning ratio of 77.8\%, ResPrune remains the best-performing approach with a relative performance of 99.3\%, demonstrating strong robustness to increased compression.
In this setting, ResPrune continues to rank first on several benchmarks, including GQA, SQA-I, POPE, TextVQA, VQA-v2, and MMB-en.
When the pruning ratio is further increased to 88.9\%, ResPrune still achieves the highest performance retention (98.0\%).
While many baselines suffer pronounced performance degradation on general VQA and LVLM-specific comprehensive benchmarks, ResPrune preserves strong performance on challenging tasks such as POPE and VizWiz, highlighting the robustness of the proposed subspace expansion with textual conditioning.

An interesting observation is that ResPrune achieves consistently strong performance on POPE, and even surpasses the full-token baseline under the most aggressive pruning ratio of 88.9\%.
We attribute this behavior to two complementary factors.
First, ResPrune suppresses visually irrelevant or weakly grounded tokens that may otherwise introduce spurious correlations, thereby encouraging more grounded and conservative responses.
Second, the textual queries in POPE are explicitly formulated as object existence verification.
After text cleaning and noun extraction, ResPrune is able to obtain a highly reliable textual guidance signal that directly corresponds to concrete visual entities.
This strong alignment between extracted nouns and discriminative visual tokens further enhances the effectiveness of text-conditioned pruning.

\subsubsection{With LLaVA-NeXT-7B} 
We further evaluate ResPrune on LLaVA-NeXT-7B, a high-resolution LVLM that produces substantially more visual tokens due to its sub-image partitioning strategy.
As shown in Table~\ref{tab:next}, ResPrune consistently outperforms all competing baselines under both moderate and aggressive pruning ratios.
At a moderate pruning ratio of 66.7\%, ResPrune achieves the highest relative performance (99.6\%), closely approaching the full-token baseline.
It delivers the best results across all reported benchmarks.
Notably, ResPrune attains particularly strong advantages on reasoning-oriented benchmarks such as GQA and SQA-I, as well as on POPE, indicating its effectiveness in preserving semantically critical visual information under high-resolution settings.
Under a more aggressive pruning ratio of 88.9\%, ResPrune continues to demonstrate clear advantages, achieving the highest overall performance retention (98.1\%).
While most baselines exhibit substantial performance degradation especially on POPE and TextVQA, ResPrune maintains strong and stable performance across nearly all tasks.
It achieves the best performance on all benchmarks except VizWiz, highlighting its robustness to severe token reduction across a broad range of tasks.
These results indicate that ResPrune scales effectively to high-resolution LVLMs with larger visual token counts.

\begin{table}[]
\centering
\caption{Ablation study of the main components in ResPrune. 
Setting-1 removes text preprocessing, Setting-2 removes textual relevance, and Setting-3 removes subspace reconstruction.
``Rel. Perf.'' denotes relative performance compared to the full-token baseline.}
\label{tab:main_comp}
\resizebox{\columnwidth}{!}{%
\begin{tabular}{lcccccc}
\toprule
\textbf{Ablation Setting} & \textbf{Rel. Perf.} & \textbf{MME} & \textbf{GQA} & \textbf{POPE} & \textbf{TextVQA} & \textbf{MMB} \\
\midrule\midrule
LLaVA-1.5-7B              & 100\%               & 1862         & 61.9         & 85.9          & 58.2             & 64.7         \\
\midrule
\rowcolor[HTML]{F5F6F7} 
\multicolumn{7}{c}{\cellcolor[HTML]{F5F6F7} Retain 128 visual tokens (pruning ratio = 77.8\%)}   \\
\midrule
\rowcolor[HTML]{F0F4FF} 
Full Method               & 98.4                & 1792         & 60.1         & 87.6          & 57.8             & 63.0         \\
Setting-1                 & 97.5                & 1754         & 59.8         & 87.5          & 57.1             & 62.6         \\
Setting-2                 & 96.6                & 1773         & 59.7         & 86.7          & 54.6             & 62.5         \\
Setting-3                 & 82.4                & 1668         & 55.6         & 83.5          & 25.4             & 59.5         \\
\midrule
\rowcolor[HTML]{F5F6F7} 
\multicolumn{7}{c}{\cellcolor[HTML]{F5F6F7} Retain 64 visual tokens (pruning ratio = 88.9\%)}    \\
\midrule
\rowcolor[HTML]{F0F4FF} 
Full Method               & 96.4                & 1740         & 58.6         & 87.5          & 57.2             & 60.7         \\
Setting-1                 & 94.7                & 1677         & 58.0         & 87.0          & 55.2             & 60.6         \\
Setting-2                 & 93.8                & 1672         & 58.0         & 86.1          & 53.5             & 60.5         \\
Setting-3                 & 79.1                & 1600         & 54.1         & 81.9          & 23.3             & 56.2        \\
\bottomrule
\end{tabular}}
\end{table}

\subsubsection{With Qwen2.5-VL-7B}
We further evaluate ResPrune on Qwen2.5-VL-7B, an advanced LVLM that supports dynamic input resolutions by unfreezing the vision encoder during multimodal finetuning.
As shown in Table~\ref{tab:qwen}, ResPrune consistently outperforms competing baselines across all pruning ratios.
At a moderate pruning ratio of 66.7\%, ResPrune achieves the highest relative performance (98.4\%), closely preserving the full-token performance. It attains the best results on MME, POPE, TextVQA, and GQA, and remains highly competitive on MMB-en.
When the pruning ratio increases to 77.8\%, ResPrune continues to demonstrate strong robustness, achieving the highest overall performance retention (96.9\%).
It consistently yields superior performance across all evaluated benchmarks, particularly on POPE and TextVQA, which require accurate grounding between visual evidence and textual queries.
Under the most aggressive pruning ratio of 88.9\%, ResPrune still maintains a clear advantage, achieving the highest relative performance (92.2\%). It again achieves the highest performance across all benchmarks, demonstrating remarkable resilience to severe token reduction.

We also observe a general trend that all pruning methods exhibit lower performance retention under the same pruning ratios when transferring from the LLaVA family to the more advanced Qwen2.5-VL model.
Compared to LLaVA family, Qwen2.5-VL family is trained with more refined multimodal training strategies and substantially larger-scale datasets, resulting in visual tokens that encode richer and more fine-grained semantic information.
As a consequence, discarding an identical proportion of visual tokens may lead to a more pronounced information loss, making token pruning intrinsically more challenging in this setting.
Despite this increased difficulty, ResPrune consistently achieves the highest performance retention across all pruning ratios.

\subsection{Ablation Analysis}

\begin{table}[]
\centering
\caption{Ablation study of the textual guidance strength $\alpha$ on LLaVA-1.5-7B and Qwen2.5-VL-7B. ``Rel. Perf.'' denotes relative performance compared to the full-token baseline.}
\label{tab:alpha}
\resizebox{\columnwidth}{!}{%
\begin{tabular}{lcccccc}
\toprule
\multicolumn{1}{c}{\textbf{Alpha}} & \textbf{Rel. Perf.} & \textbf{MME} & \textbf{GQA} & \textbf{POPE} & \textbf{TextVQA} & \textbf{MMB} \\
\midrule
\midrule
LLaVA-1.5-7B                       & 100\%               & 1862         & 61.9         & 85.9          & 58.2             & 64.7         \\
\midrule
\rowcolor[HTML]{F5F6F7} 
\multicolumn{7}{c}{\cellcolor[HTML]{F5F6F7} Pruning ratio = 77.8\%}                                       \\
\midrule
Alpha = 0.25                       & 97.7                & 1752         & 60.1         & 87.7          & 57.4             & 62.4         \\
Alpha = 0.5                        & 98.1                & 1778         & 59.9         & 87.7          & 57.9             & 62.7         \\
\rowcolor[HTML]{F0F4FF} 
Alpha = 0.75                       & 98.4                & 1792         & 60.1         & 87.6          & 57.8             & 63.0         \\
Alpha = 1.0                        & 98.3                & 1801         & 60.0         & 87.5          & 57.7             & 62.6         \\
Alpha = 1.25                       & 98.1                & 1789         & 59.8         & 87.5          & 57.7             & 62.6         \\
\midrule
\midrule
Qwen2.5-VL-7B                      & 100\%               & 2321         & 60.8         & 87.1          & 85.2             & 83.2         \\
\midrule

\rowcolor[HTML]{F5F6F7} 
\multicolumn{7}{c}{\cellcolor[HTML]{F5F6F7} Pruning ratio = 77.8\%}                                      \\
\midrule
Alpha = 0.1                        & 96.3                & 2257         & 59.0         & 86.2          & 77.3             & 81.3         \\
Alpha = 0.2                        & 96.5                & 2255         & 59.3         & 86.9          & 76.8             & 81.6         \\
\rowcolor[HTML]{F0F4FF} 
Alpha = 0.3                        & 96.9                & 2246         & 59.4         & 87.3          & 78.0             & 81.7         \\
Alpha = 0.4                        & 96.7                & 2253         & 59.5         & 87.2          & 76.8             & 81.9         \\
Alpha = 0.5                        & 96.4                & 2239         & 59.3         & 87.2          & 76.3             & 81.8    \\
\bottomrule
\end{tabular}}
\end{table}

This section systematically analyzes how different components, design choices and hyperparameters influence the effectiveness of ResPrune.

\subsubsection{Contribution of Main Components}

We analyze the contribution of the main components in ResPrune by ablating individual design choices while keeping all other settings unchanged.
Table~\ref{tab:main_comp} reports results on LLaVA-1.5-7B under two pruning ratios of 77.8\% and 88.9\%.

Setting-1 removes the text preprocessing step (i.e., instruction format removal and noun extraction) while keeping the residual energy–based token selection and textual guidance unchanged.
Compared to the full method, this variant results in a consistent but relatively mild performance drop across most benchmarks. 
The degradation is more noticeable on text-centric tasks such as TextVQA, indicating that text preprocessing helps produce cleaner and more reliable textual guidance signals, but is not the dominant factor for overall performance.

Setting-2 removes textual guidance entirely and performs iterative token selection solely based on the plain residual energy scores.
This variant leads to a more pronounced performance decrease than setting-1, particularly on instruction-sensitive benchmarks such as POPE and TextVQA.
These results suggest that incorporating textual relevance is critical for aligning token selection with the task instruction, especially when aggressive pruning ration is applied.

Setting-3 removes the progressive token selection process and retains visual tokens via a single-pass top-$k$ ranking based solely on textual relevance scores.
This variant exhibits a substantial performance collapse under both pruning ratios, with especially severe degradation on TextVQA and noticeable drops across all benchmarks.
This suggests that textual relevance alone is insufficient for effective visual token pruning, as it fails to preserve the representational coverage of the visual embedding space.

These ablation results together confirm that subspace reconstruction forms the core of ResPrune, ensuring broad and non-redundant visual coverage, while textual guidance acts as a crucial complementary signal that aligns token selection with task semantics. Text preprocessing further refines this guidance by reducing noise, but plays a secondary role compared to the other two components.

\subsubsection{Strength of Textual Guidance}

We next investigate the effect of the textual guidance strength $\alpha$ on the performance of ResPrune.
Table~\ref{tab:alpha} reports results on LLaVA-1.5-7B and Qwen2.5-VL-7B.
We include Qwen2.5-VL-7B in this analysis to account for architectural differences in the vision encoder compared to the LLaVA family, which may affect the optimal strength of textual guidance.

On LLaVA-1.5-7B, as $\alpha$ increases from 0.25 to 0.75, the overall performance retention improves steadily, reaching its peak at $\alpha=0.75$.
Further increasing $\alpha$ leads to marginal performance fluctuations, but no severe degradation is observed. 
This suggests that moderate textual guidance effectively biases token selection toward instruction-relevant content while still preserving visual coverage through residual energy modeling.

A similar trend is observed on Qwen2.5-VL-7B, although the optimal $\alpha$ is smaller.
Performance improves as $\alpha$ increases from 0.1 to 0.3, reaching the best performance retention at $\alpha = 0.3$. Beyond this point, stronger guidance slightly degrades performance. This behavior indicates that, due to the higher semantic density of visual tokens in Qwen2.5-VL, overly strong textual bias may lead to excessive focus on instruction-related regions at the expense of global coverage.

These results demonstrate that ResPrune is relatively insensitive to the exact choice of $\alpha$ within a reasonable range, while an intermediate guidance strength consistently yields the best trade-off between visual coverage and instruction alignment.

\subsubsection{Formulation of Textual Guidance}
\label{sec:text_formulation}

\begin{table}[]
\centering
\caption{Ablation study on different formulations of textual relevance computation.
Details of each setting are described in Section~\ref{sec:text_formulation}.
``Rel. Perf.'' denotes the average relative performance with respect to the full-token baseline.}
\label{tab:text_computation}
\resizebox{\columnwidth}{!}{%
\begin{tabular}{lcccccc}
\toprule
\textbf{Setting}  & \textbf{Rel. Perf.}  & \textbf{MME}  & \textbf{GQA}  & \textbf{POPE}  & \textbf{TextVQA}  & \textbf{MMB}  \\
\midrule
\midrule
LLaVA-1.5-7B      & 100\%                & 1862          & 61.9          & 85.9           & 58.2              & 64.7          \\
\midrule

\rowcolor[HTML]{F5F6F7} 
\multicolumn{7}{c}{\cellcolor[HTML]{F5F6F7} Retain 128 visual tokens (pruning ratio = 77.8\%)} \\
\midrule
\rowcolor[HTML]{F0F4FF} 
Setting-1         & 98.4                 & 1792          & 60.1          & 87.6           & 57.8              & 63.0          \\
Setting-2         & 97.8                 & 1778          & 59.9          & 87.7           & 56.6              & 62.9          \\
Setting-3         & 95.4                 & 1775          & 59.6          & 87.5           & 51.2              & 61.8         \\
\midrule
\end{tabular}}
\end{table}

We investigate alternative formulations of textual relevance in ResPrune, aiming to understand how different aggregation strategies over textual tokens affect pruning performance.
Table~\ref{tab:text_computation} reports results on LLaVA-1.5-7B under a fixed pruning ratio of 77.8\%.

Setting-1 computes textual relevance as the maximum cosine similarity between a visual token and all textual tokens (as defined in Eq.~\ref{eq:text_relevance}). Setting-2 computes textual relevance as the average similarity between a visual token and all textual tokens.
Setting-3 first aggregates all textual tokens into a single embedding via average pooling and then computes similarity with visual tokens. As shown in Table~\ref{tab:text_computation}, Setting-1 achieves the best overall relative performance (98.4\%), outperforming the average-based formulation in Setting-2 (97.8\%) and the pooled-text formulation in Setting-3 (95.4\%).

We attribute the superiority of the max-based formulation to its robustness against noisy or weakly relevant textual tokens. In LVLM prompts, even after text preprocessing, the remaining textual tokens may still include modifiers, function words, or loosely related concepts that are not directly grounded in the image. Averaging similarities over all tokens implicitly treats these tokens as equally informative, which can dilute the guidance signal from the truly discriminative keywords. The pooled-text formulation (Setting-3) further exacerbates this issue by collapsing all textual tokens into a single embedding before computing relevance. In contrast, the max operator acts as a selective mechanism that emphasizes the strongest visual-text correspondence and suppresses the influence of irrelevant or noisy text tokens.

These results support our design choice of using a max-over-token relevance formulation, as it provides a sharper and more noise-robust guidance signal for text-conditioned pruning.

\subsubsection{Seed Token Selection}
\label{sec:seed_selection}

\begin{table}[]
\centering
\caption{Ablation study on different seed token selection strategies. Details of each setting are described in Section~\ref{sec:seed_selection}. ``Rel. Perf.'' denotes the average relative performance with respect to the full-token baseline.}
\label{tab:token_selection}
\resizebox{\columnwidth}{!}{%
\begin{tabular}{lcccccc}
\toprule
\textbf{Setting} & \textbf{Rel. Perf.} & \textbf{MME} & \textbf{GQA} & \textbf{POPE} & \textbf{TextVQA} & \textbf{MMB} \\
\midrule
\midrule
LLaVA-1.5-7B     & 100\%               & 1862         & 61.9         & 85.9          & 58.2             & 64.7         \\
\midrule
\rowcolor[HTML]{F5F6F7} 
\multicolumn{7}{c}{\cellcolor[HTML]{F5F6F7} Pruning ratio = 77.8\%, CLS token present}  \\
\midrule
\rowcolor[HTML]{F0F4FF} 
Setting-1        & 98.4                & 1792         & 60.1         & 87.6          & 57.8             & 63.0         \\
Setting-2        & 98.0                & 1773         & 60.0         & 87.3          & 57.7             & 62.7         \\
Setting-3        & 97.7                & 1755         & 59.9         & 87.8          & 57.7             & 62.1         \\
Setting-4        & 97.7                & 1753         & 59.9         & 87.5          & 57.9             & 62.2         \\
Setting-5        & 97.6                & 1753         & 59.9         & 87.7          & 57.4             & 62.4         \\
\midrule\midrule
Qwen2.5-VL-7B    & 100\%               & 2321         & 60.8         & 87.1          & 85.2             & 83.2         \\
\midrule

\rowcolor[HTML]{F5F6F7} 
\multicolumn{7}{c}{\cellcolor[HTML]{F5F6F7} Pruning ratio = 77.8\%, CLS token absent}  \\
\midrule
\rowcolor[HTML]{F0F4FF} 
Setting-2        & 96.9                & 2246         & 59.4         & 87.3          & 78.0             & 81.7         \\
Setting-3        & 96.6                & 2236         & 59.5         & 87.2          & 76.7             & 82.2         \\
Setting-4        & 96.5                & 2252         & 59.6         & 87.1          & 76.9             & 80.8         \\
Setting-5        & 96.3                & 2215         & 59.6         & 87.1          & 76.8             & 81.4       \\
\bottomrule
\end{tabular}}
\end{table}

We study the impact of different seed token selection strategies in ResPrune under a fixed pruning ratio of 77.8\%, evaluated on both LLaVA-1.5-7B and Qwen2.5-VL-7B. The results are summarized in Table~\ref{tab:token_selection}.

As the seed token defines the initial direction of subspace expansion, its choice may influence the subsequent greedy selection process.
We consider five representative initialization strategies.
Setting-1 selects the visual token receiving the highest attention weight from the CLS token, which is only available in the LLaVA family.
Setting-2 selects the token with the largest $\ell_2$ norm, corresponding to the highest feature magnitude. 
Setting-3 chooses the token with the highest textual relevance score, i.e., the token most aligned with the cleaned textual instruction.
Setting-4 selects the token that is most similar to the global average-pooled visual embedding, aiming to capture a globally representative visual direction.
Setting-5 selects the token closest to the spatial center of the 2D token grid, based on the heuristic that central regions are more likely to overlap with salient content.

\begin{table*}[t]
\centering
\caption{Practical efficiency analysis of ResPrune on LLaVA-NeXT-7B measured on the MME benchmark. Relative changes with respect to the full-token baseline are shown in parentheses and highlighted in green.}
\label{tab:efficiency}
\small
\setlength{\tabcolsep}{4pt}
\renewcommand{\arraystretch}{1.15}
\resizebox{\textwidth}{!}{%
\begin{tabular}{lcccccc}
\toprule
\textbf{Method} &
\textbf{FLOPs (T)} &
\textbf{KV-cache (MB)} &
\textbf{GPU Memory (GB)} &
\textbf{Throughput (tokens/sec.)} &
\textbf{Evaluation Time (min.:sec.)} &
\textbf{Performance} \\
\midrule
\midrule
\rowcolor[HTML]{F5F6F7}
\multicolumn{7}{c}{Retain up to 2880 visual tokens} \\
LLaVA-NeXT-7B & 30.6 & 1084.7 & 19.8 & 16.1 & 9:05 & 1842 \\
\addlinespace
\midrule
\rowcolor[HTML]{F5F6F7}
\multicolumn{7}{c}{Retain up to 640 visual tokens (pruning ratio = 66.7\%)} \\
ResPrune
& 9.6  \textcolor{green!60!black}{($\downarrow 68.6\%$)}
& 361.2 \textcolor{green!60!black}{($\downarrow 66.7\%$)}
& 17.8 \textcolor{green!60!black}{($\downarrow 10.1\%$)}
& 33.5 \textcolor{green!60!black}{($2.08\times$)}
& 4:39 \textcolor{green!60!black}{($\downarrow 48.8\%$)}
& 1821 \textcolor{green!60!black}{($\downarrow 1.1\%$)} \\
\addlinespace
\midrule
\rowcolor[HTML]{F5F6F7}
\multicolumn{7}{c}{Retain up to 320 visual tokens (pruning ratio = 88.9\%)} \\
ResPrune
& 3.1  \textcolor{green!60!black}{($\downarrow 89.9\%$)}
& 120.4 \textcolor{green!60!black}{($\downarrow 88.9\%$)}
& 17.7 \textcolor{green!60!black}{($\downarrow 10.6\%$)}
& 36.7 \textcolor{green!60!black}{($2.28\times$)}
& 3:28 \textcolor{green!60!black}{($\downarrow 61.8\%$)}
& 1780 \textcolor{green!60!black}{($\downarrow 3.3\%$)} \\
\bottomrule
\end{tabular}%
}
\end{table*}

\begin{figure*}[t]
  \centering
  \includegraphics[width=\textwidth]{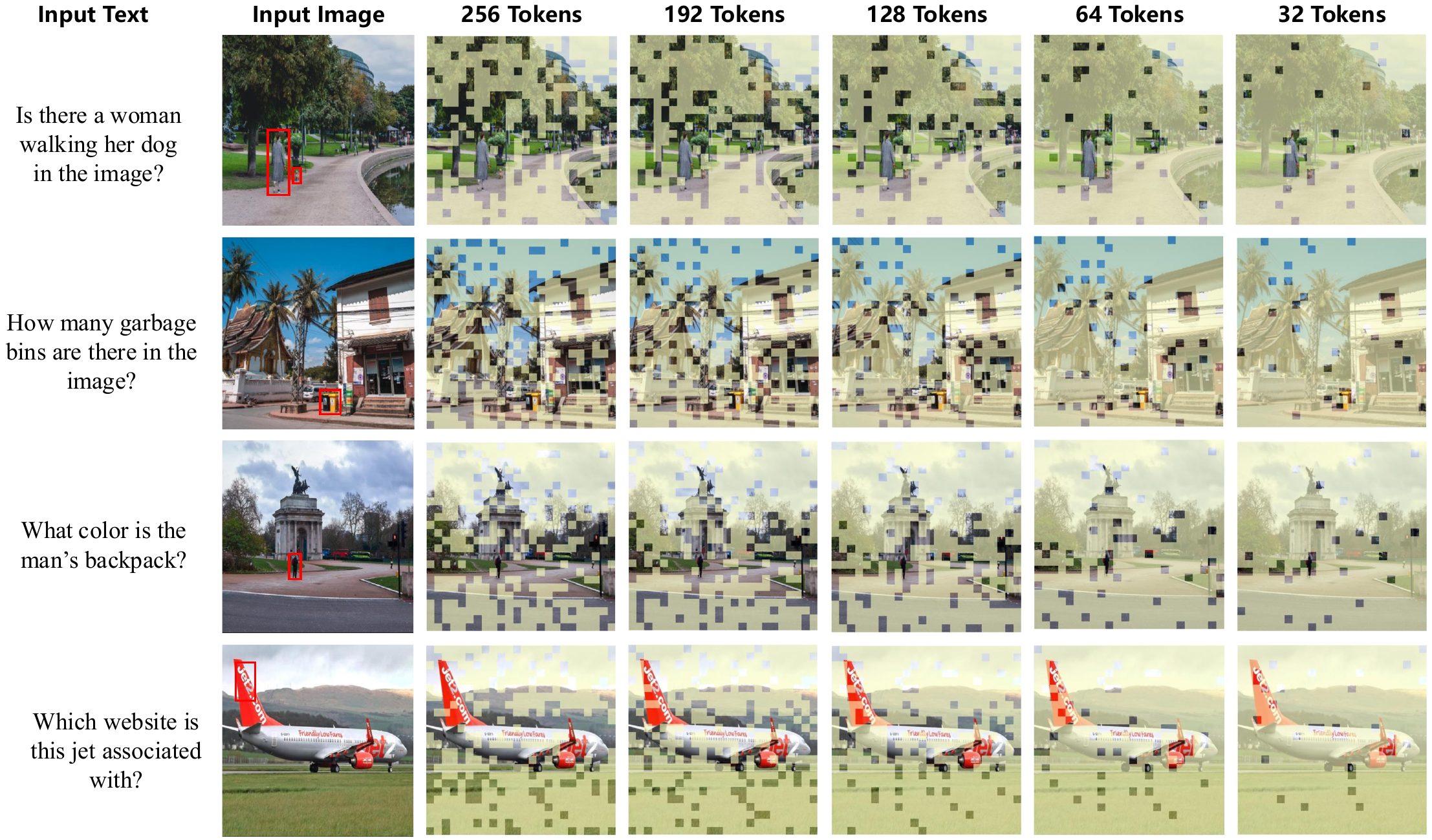}
  \caption{Qualitative visualization of ResPrune’s pruning results on LLaVA-1.5-7B under different visual token budgets. Masked patches correspond to pruned visual tokens, while unmasked patches indicate retained tokens. Answer-relevant regions are highlighted with red bounding boxes.}
  \label{fig:case_study}
\end{figure*}

On LLaVA-1.5-7B, where a CLS token is available, Setting-1 yields the best overall performance.
This strategy consistently outperforms alternative initializations based on token norm, textual relevance, global pooled similarity, or spatial centrality.
These results suggest that CLS attention provides a reliable global saliency signal, serving as an effective and informative starting point for subspace construction.

On Qwen2.5-VL-7B, where the CLS token is absent, selecting the visual token with the largest $\ell_2$ norm (Setting-2) achieves the best performance.
Other strategies, including text-driven and geometry-based initialization, lead to slightly inferior results.
This observation indicates that, in the absence of explicit global attention signals, selecting a dominant and high-energy token offers a stable and architecture-agnostic initialization for ResPrune.

Overall, while an appropriate seed token selection can further improve performance, ResPrune remains robust to different initialization strategies and does not rely critically on a specific choice of seed token.

\subsection{Practical Efficiency Analysis}

We further evaluate the practical efficiency of ResPrune by measuring its end-to-end inference cost on LLaVA-NeXT-7B using the MME benchmark.
Following common practice, we report system-level metrics including FLOPs, KV-cache memory, GPU memory footprint, decoding throughput, and total evaluation time.
All experiments are conducted using four NVIDIA L20 GPUs with a batch size of 1, and the results are summarized in Table~\ref{tab:efficiency}.

When applying ResPrune with a moderate pruning ratio of 66.7\%, the prefill FLOPs drop from 30.6T to 9.6T, and KV-cache memory is reduced by approximately 67\%.
This leads to a more than twofold increase in decoding throughput to 33.5 tokens per second, and reduces the overall evaluation time from 9:05 to 4:39 minutes, while maintaining nearly full performance.
Under a more aggressive pruning ratio of 88.9\%, these efficiency gains are further amplified.
The FLOPs are reduced by nearly an order of magnitude to 3.1T, KV-cache memory drops below 130~MB, and throughput increases to 36.7 tokens per second.
Consequently, the evaluation time is shortened to 3:28 minutes, corresponding to a 62\% reduction compared to the full-token baseline.

These results demonstrate that ResPrune provides significant practical efficiency improvements in real world scenarios.
By shifting computation from the quadratic-cost LLM self-attention to a lightweight, training-free token selection stage, ResPrune enables fast and memory-efficient inference while largely preserving model accuracy.

\subsection{Case Study}
\label{sec:case_study}

We provide qualitative case studies in Figure~\ref{fig:case_study} to illustrate how ResPrune selects visual tokens on LLaVA-1.5-7B under different token budgets. The selected examples cover diverse tasks, including object existence verification, object counting, attribute recognition, and text-centric reasoning.

Across all cases, ResPrune exhibits consistent and interpretable pruning behavior.
With relatively high token budgets (e.g., 256 or 192 tokens), the retained tokens are broadly distributed across the image, preserving both global contextual information and discriminative fine-grained details. As the token budget decreases, visually redundant regions such as background vegetation, sky, or uniform textures are progressively suppressed, while the retained tokens increasingly concentrate on instruction-relevant regions.

Overall, these qualitative results corroborate our quantitative findings.
By jointly considering residual energy and textual relevance, ResPrune achieves a favorable balance between visual coverage and task awareness, enabling LVLMs to preserve critical visual evidence even under aggressive token reduction.
 
\section{Conclusion}
In this work, we presented ResPrune, a training-free visual token pruning framework for efficient inference in LVLMs. ResPrune formulates visual token pruning as a subspace reconstruction problem and adopts a greedy solution based on residual energy to progressively select informative visual tokens.
By further conditioning the selection process on textual relevance, ResPrune effectively balances visual coverage and instruction awareness, enabling the construction of compact yet semantically meaningful token subset.
Extensive experiments across multiple LVLM backbones demonstrate both the effectiveness and the model-agnostic nature of ResPrune.

Despite its effectiveness, limitations may exist. For example, the strength of textual guidance is currently controlled by a fixed hyperparameter $\alpha$, which is manually tuned for different model families and remains constant across inputs. This design may not fully account for variations in instruction specificity or visual complexity across different samples.
An important direction for future work is therefore to explore adaptive or learning-based mechanisms that dynamically adjust the strength of textual guidance based on input characteristics or model feedback.

\bibliographystyle{IEEEtran}
\bibliography{references}

\vfill

\end{document}